\documentclass{article} 
\usepackage{iclr2027_conference,times}

\usepackage[utf8]{inputenc}
\usepackage[T1]{fontenc}
\usepackage{hyperref}
\usepackage{url}
\usepackage{booktabs}
\usepackage{amsfonts}
\usepackage{amsmath}
\usepackage{amssymb}
\usepackage{nicefrac}
\usepackage{microtype}
\usepackage{xcolor}
\usepackage{graphicx}
\usepackage{subcaption}
\usepackage{calc}
\usepackage{tikz}
\usepackage{pgfplots}
\usepackage{enumitem}
\usepackage{float}   
\usepackage{amsthm}
\usepackage{orcidlink}

\newtheorem{proposition}{Proposition}

\pgfplotsset{compat=1.18}

\AtBeginDocument{%
  \setlength{\abovedisplayskip}{2pt plus 1pt minus 1pt}%
  \setlength{\belowdisplayskip}{2pt plus 1pt minus 1pt}%
  \setlength{\abovedisplayshortskip}{2pt plus 1pt minus 1pt}%
  \setlength{\belowdisplayshortskip}{2pt plus 1pt minus 1pt}%
}

\setlist{topsep=2pt, partopsep=0pt, parsep=2pt, itemsep=2pt}

\makeatletter
\renewcommand{\paragraph}{%
  \@startsection{paragraph}{4}{\z@}%
                {0.3ex plus 0.1ex minus 0.05ex}%
                {-1em}%
                {\normalsize\bf}%
}
\makeatother

\title{Federation of Experts: Communication Efficient Distributed Inference for Large Language Models}

\author{%
  Muhammad Shahir Abdurrahman~\orcidlink{0009-0006-5723-9997} \\
  Stanford University \\
  \texttt{shahir@stanford.edu} \\
  \And
  Chun Deng \\
  Stanford University \\
  \texttt{dengchun@stanford.edu} \\
  \And
  Azalia Mirhoseini \\
  Stanford University \\
  \texttt{azalia@stanford.edu} \\
  \And
  Philip Levis \\
  Stanford University \\
  \texttt{plevis@stanford.edu} \\
}

\newcommand{\FwdSpeedupFlex}{5.88}  
\newcommand{\FwdSpeedupVLLM}{5.58}  
\newcommand{\FwdBatchMax}{24{,}576}

\newcommand{\TTFTSpeedupSingle}{2.75}  
\newcommand{\TTFTSpeedupVLLM}{3.63}  

\newcommand{\TTFTSpeedupBest}{3.66}  
\newcommand{\TPOTSpeedupBest}{1.53}  
\newcommand{\TTFTSpeedupSingleAtHalf}{2.72}  
\newcommand{\TTFTSpeedupMultiAtHalf}{2.90}  
\newcommand{\TailTTFTFoEMulti}{0.872}  
\newcommand{\TailTTFTMoEMulti}{3.375}  

\newcommand{\CommReduction}{12.3}  

\newcommand{\LoadDiscrepancyMoE}{1.44}  
\newcommand{\LARSemMoE}{0.39}  
\newcommand{\LARMoETuner}{0.13}  

\newcommand{\FitCoefMoE}{2.1 \times 10^{-3}}  
\newcommand{\FitCoefFoE}{3 \times 10^{-4}}  

\iclrfinalcopy  
\begin{document}

\maketitle

\lhead{}

\begin{abstract}

Mixture of experts has emerged as the primary mechanism for making Large Language Models (LLMs) computationally efficient. However, in distributed settings, communicating token embeddings between experts is a significant bottleneck.

We present the novel Federation of Experts (FoE) architecture. FoE restructures the MoE block of a transformer layer into multiple MoE clusters. Each cluster is responsible for only one of the KV heads and expert parallelism is applied between those experts. Between clusters, a sum synchronizes the post-attention residuals, which then drives routing and dispatch for the next MoE block. In a single-node setting, FoE completely eliminates all-to-all communication as all experts within a group are contained on the same GPU. In multi-node settings, FoE confines all-to-all communication to the intra-node fabric, thus significantly reducing communication overhead.

An implementation of FoE finds that on LongBench, FoE significantly improves inference throughput and latency in both single-node and multi-node settings, reducing end-to-end prefill latency by up to \FwdSpeedupFlex$\times$, TTFT by \TTFTSpeedupBest$\times$, and TPOT by \TPOTSpeedupBest$\times$. It does so while achieving comparable generation quality to a mixture of experts model of the same size and training configuration.

\end{abstract}

\section{Introduction}

Large Language Models (LLMs) are the backbone for many modern AI applications. Driven by the transformer architecture \citep{vaswani2023attentionneed}, these models consistently improve as parameter counts increase~\citep{brown2020languagemodelsfewshotlearners,touvron2023llamaopenefficientfoundation}. However, computation scales proportionally with parameters, imposing limits on both training efficiency and inference deployment.

Mixture-of-Experts (MoE) architectures~\citep{Shazeer2017OutrageouslyLN} mitigate this by activating only a subset of experts per token, decoupling total parameters from active parameters, and have been validated to hundreds of billions of parameters~\citep{fedus2022switchtransformersscalingtrillion,jiang2024mixtralexperts}. These compute efficiency gains, however, introduce a new bottleneck in distributed settings: the sparse routing mechanism in expert parallelism. When a token is assigned to an expert residing on a different GPU, it must be dispatched across the network, relying on expensive bandwidth-bounded communication primitives. This worsens with the number of GPUs, and accounts for 68\%+ of end-to-end latency \citep{flashmoe}. For example, in DeepSeek production deployments for their V3/R1 models, 256 experts are distributed across 18 nodes for decode and 4 nodes for prefill, with 8 GPUs per node, in order to achieve higher throughput with larger batch sizes \citep{deepseek2024v3}. With expansive intra-node GPU interconnect fabrics and intra-node networking becoming increasingly common, the communication overhead is a very important problem to address.

\begin{figure}[htbp]
    \centering
    \begin{subfigure}[c]{\widthof{\includegraphics[scale=0.4]{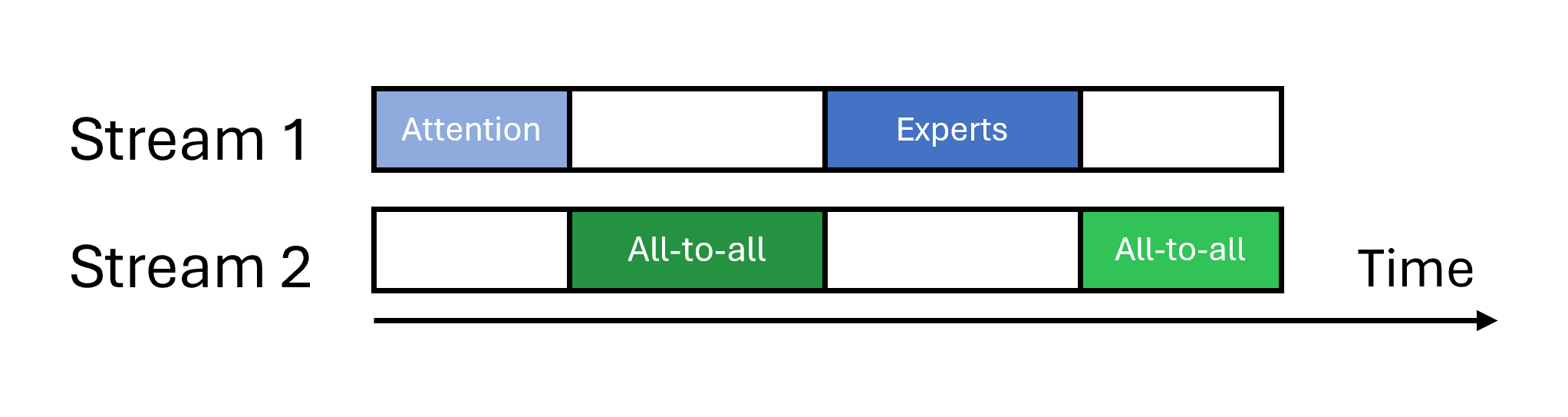}}}
        \centering
        \includegraphics[scale=0.4]{assets/moe-timing.png}
        \caption{Mixture of Experts (MoE)}
        \label{fig:moe-timing}
    \end{subfigure}\hspace{1em}%
    \begin{subfigure}[c]{\widthof{\includegraphics[scale=0.4]{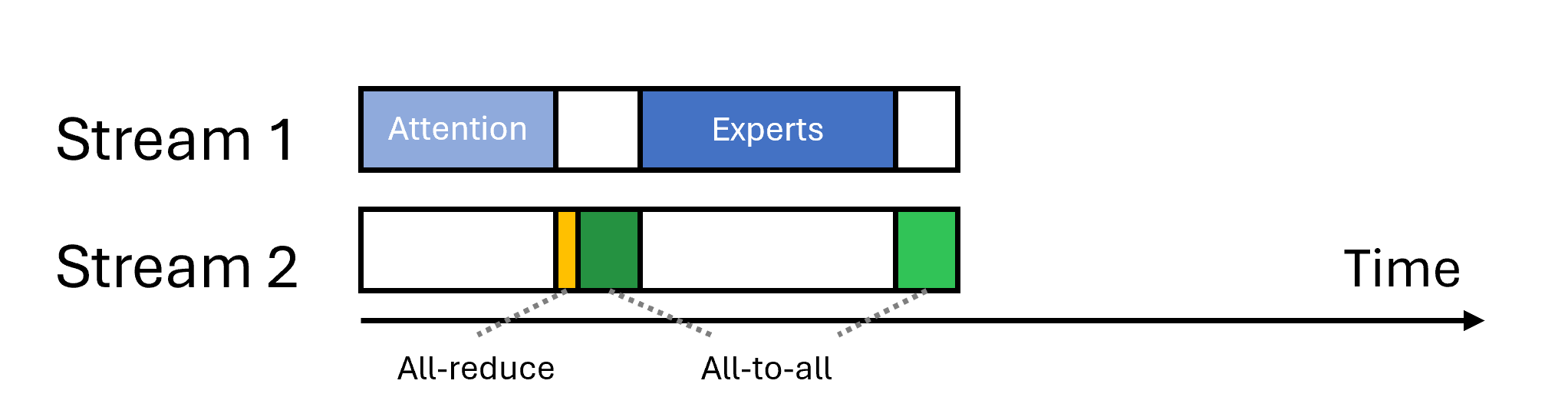}}}
        \centering
        \includegraphics[scale=0.4]{assets/foe-timing.png}
        \caption{Federation of Experts (FoE)}
        \label{fig:foe-timing}
    \end{subfigure}
    \caption{Timing diagrams comparing standard MoE and FoE. FoE confines all-to-all communication into localized clusters, mitigating the network bottleneck in distributed settings.}
    \label{fig:communication-bottleneck}
\end{figure}

When exploring communication in the MoE architecture, we find that the required all-to-all primitive for standard expert parallelism across a cluster is a major bottleneck for distributed inference, as illustrated in Figure~\ref{fig:communication-bottleneck}.

In multi-node deployments, the problem becomes more pronounced, as intra-node NVLink delivers up to 900 GB/s \citep{nvlink}, while inter-node InfiniBand caps at 400 Gb/s per link \citep{infiniband}, a gap of roughly $18\times$, the bandwidth ratio $r$ used throughout our analysis. Improvements must focus on maximizing the Local Activation Rate (LAR) - the proportion of expert selections that are resolved locally without network transfer. In standard MoE, as the hardware cluster size grows, the LAR falls, forcing the majority of tokens to leave their resident GPUs and incur substantial network latencies.

This paper introduces the Federation of Experts (FoE) architecture to reduce this communication bottleneck. FoE partitions each transformer layer into $H$ independent expert groups, each owning $1/H$ of the KV heads and $1/H$ of the experts and running on its own subset of GPUs. Every token reaches every group, selecting $k/H$ of its $k$ experts within each, and the groups synchronize once per layer through a lightweight cross-group all-reduce. All-to-all communication therefore stays inside a group: the expensive $k$-proportional traffic never crosses a group boundary, and per-group load is uniform by construction. 

This paper makes three contributions:
\begin{enumerate}
    \item The Federation of Experts (FoE) architecture, which partitions experts and KV heads into isolated groups and confines expert-parallel communication within them while guaranteeing even load balancing across expert groups.
    \item Analysis of FoE's communication complexity and gradient structure, deriving an order-of-magnitude communication reduction and showing that expressivity is preserved.
    \item Experimental results on the LongBench dataset \citep{longbench}, showing that FoE reduces end-to-end prefill latency by up to $\FwdSpeedupFlex\times$, time-to-first-token (TTFT) by $\TTFTSpeedupBest\times$, and time-per-output-token (TPOT) by $\TPOTSpeedupBest\times$, while achieving comparable generation quality to an MoE model of the same size and training configuration.
\end{enumerate}

\section{Background}

\textbf{Mixture-of-Experts.}
Mixture of Experts (MoE) models~\citep{Shazeer2017OutrageouslyLN} increase a model's total parameter count without a proportional increase in computation: rather than activating every parameter for every token as a dense model does, an MoE activates only a subset of the expert networks in each transformer block~\citep{lepikhin2020gshard, fedus2022switchtransformersscalingtrillion}. This sharply reduces active parameters per token and is now standard in open production-scale LLMs~\citep{jiang2024mixtralexperts, deepseek2024v3, olmoe}.

Because the FFN block is the largest component of a transformer layer by parameter count, it is the primary target for MoE treatment. In a token-choice MoE FFN block~\citep{Shazeer2017OutrageouslyLN, lepikhin2020gshard} (the standard formulation), the dense FFN is replaced by many expert FFNs and a gate network produces a weight distribution over them for each token. The experts with the top-$k$ weights are selected and their outputs combined in a weighted sum, so only those $k$ experts are activated for that token.

\textbf{MoE Inference Systems and Parallelism.}
Integrated serving engines such as vLLM, SGLang, and DeepSpeed-MII \citep{kwon2023vllm, zheng2024sglang, holmes2024deepspeed} optimize LLM inference through continuous batching, high-performance kernels, and graph-level techniques such as pipeline parallelism. On top of these, MoE-specialized engines \citep{rajbhandari2022deepspeed, singh2023}, and now vLLM and SGLang as well, rely heavily on Expert Parallelism (EP), which distributes experts across devices and sends each token's activations to the device holding its selected expert through all-to-all communication. As cluster size grows, the probability that a selected expert is co-resident with its token collapses, and communication becomes a dominant bottleneck, accounting for 47\% to 68\% of inference time \citep{comet, flashmoe}.

\textbf{Expert Parallelism Optimizations.}
Real-world MoE deployments suffer from unbalanced routing and network-bound hot-spotting, which prior work mitigates through expert replication and load balancing such as with EPLB and Lina \citep{deepseekeplb2024, lina, deepseek2024v3}. To mask the steep all-to-all overhead, systems-level work pipelines execution aggressively: COMET, FlashMoE, and DeepEP \citep{comet, flashmoe, deepep2025} respectively overlap computation and communication, fuse the distributed MoE operator to eliminate launch overhead, and exploit pure RDMA overlapping. Scheduling frameworks instead use expert affinity to minimize cross-node routing: Sem-MoE and GRACE-MoE \citep{semmoe, gracemoe} proactively collocate experts with their activating tokens, while ExFlow and MoETuner \citep{exflow, moetuner} exploit predictable inter-layer routing dependencies to optimize static expert-to-GPU assignment. A third line restructures the architecture itself: BigMac \citep{bigmac} projects communication into a lower dimension, Pangu Pro MoE \citep{tang2025pangupromoe} divides the router into grouped top-$k$ selection, and LatentMoE \citep{latentmoe} moves routing and computation into a compact latent space. Device-limited routing of DeepSeek-V2 and V3 \citep{deepseekv2, deepseek2024v3} caps the number of devices a token may reach, lowering the multiplier of all-to-all traffic, but is not able to reduce it significantly or deterministically and still pays inter-node traffic penalties.

These methods optimize the global dispatch process but still operate within a single unified EP group, so they remain bounded by unavoidable all-to-all communication. FoE instead takes a structural approach: restructuring global routing into disjoint, balanced expert groups natively confines traffic, guaranteeing a balanced and high Local Activation Rate (LAR) across cluster nodes and sidestepping the global all-to-all bottleneck without dynamic scheduling or low-dimensional approximation.

\section{Federation of Experts}
\label{sec:design}

\begin{figure}[b]
  \centering
  \begin{subfigure}[c]{\widthof{\includegraphics[scale=0.25]{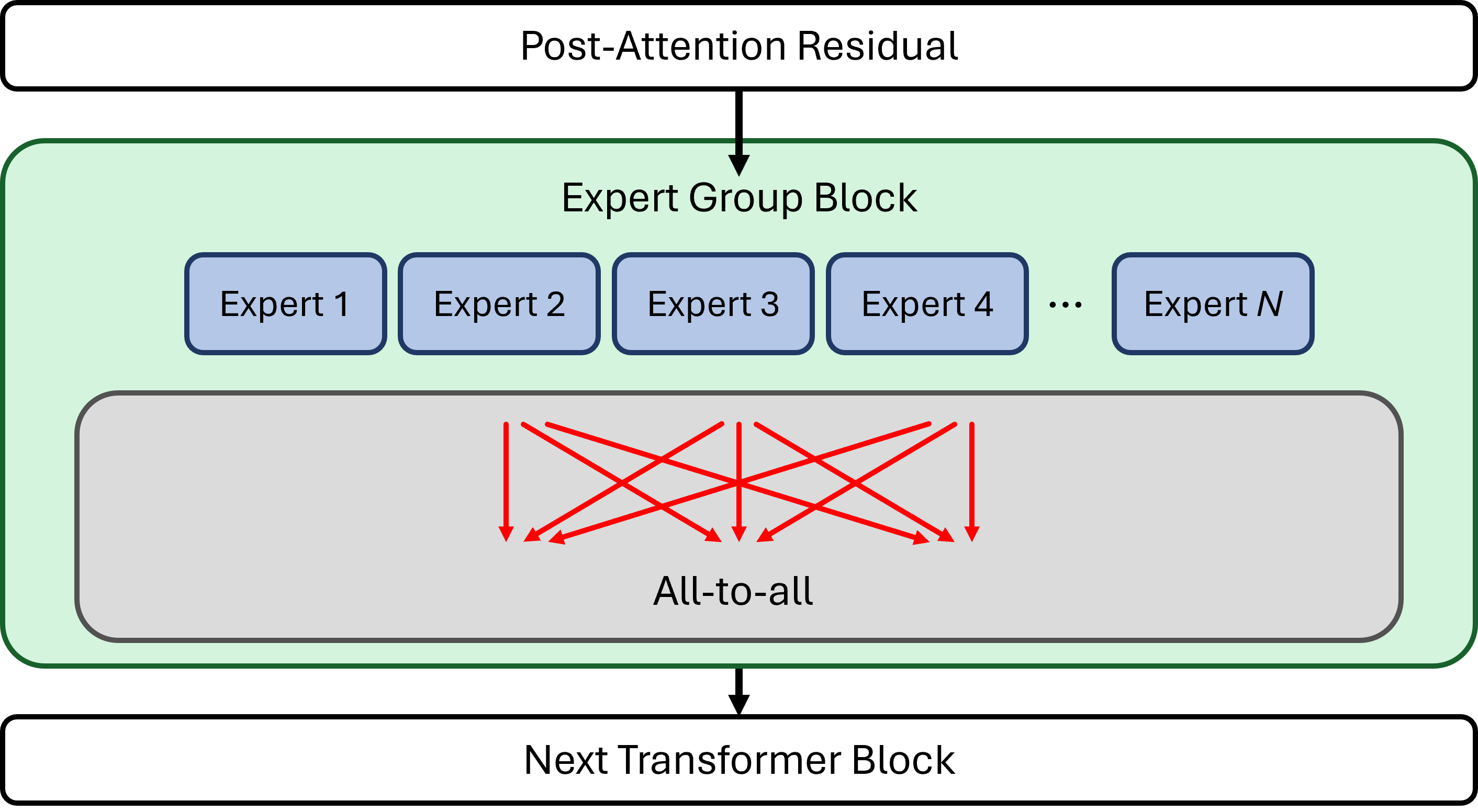}}}
    \centering
    \includegraphics[scale=0.25]{assets/moe-diagram.png}
    \caption{Mixture of Experts.}\label{fig:moe-diagram}
  \end{subfigure}\hspace{1em}%
  \begin{subfigure}[c]{\widthof{\includegraphics[scale=0.25]{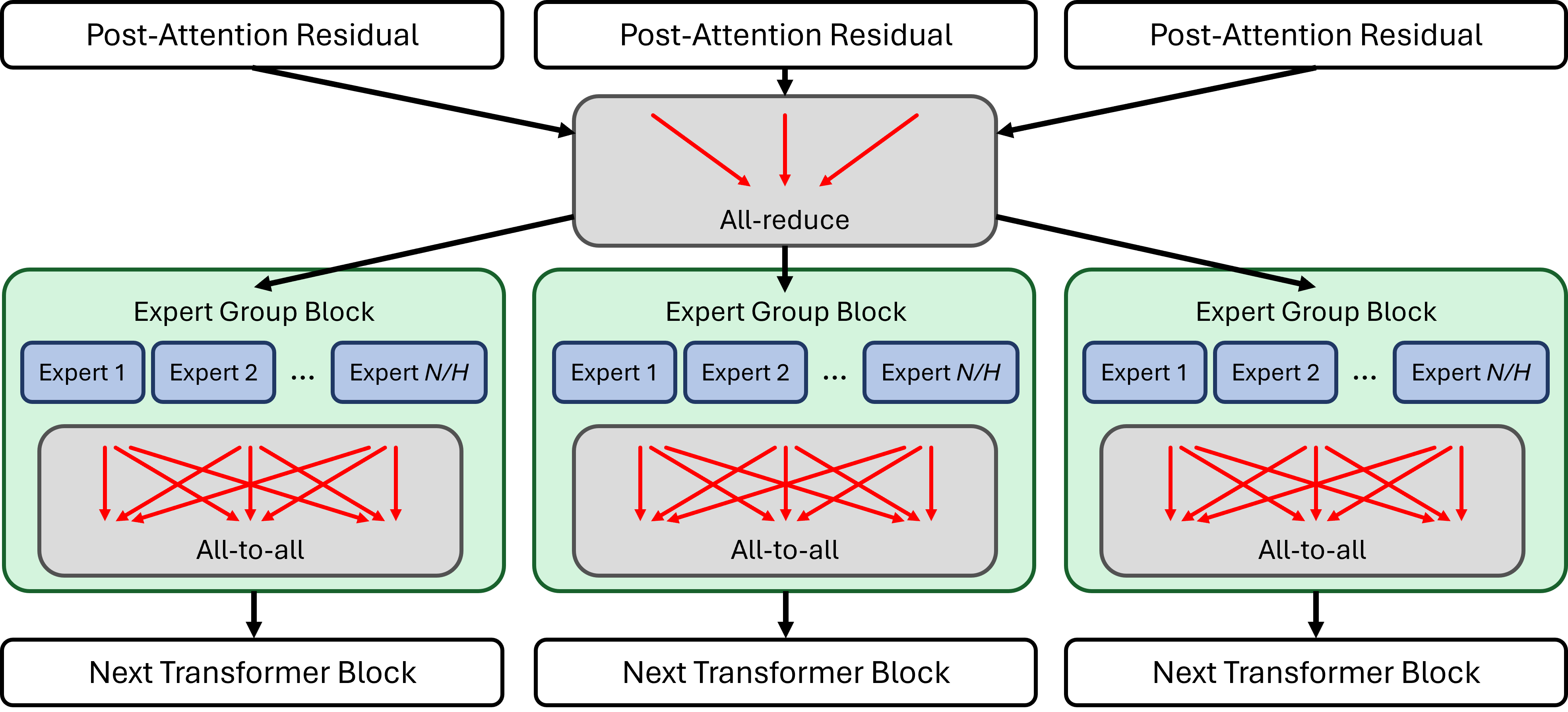}}}
    \centering
    \includegraphics[scale=0.25]{assets/foe-diagram.png}
    \caption{Federation of Experts.}\label{fig:foe-diagram}
  \end{subfigure}
  \caption{Comparison in expert communication between MoE and FoE.}\label{fig:moe-vs-foe}
\end{figure}

When experts are distributed across GPUs, a token's selected experts can reside on any of the $G$ GPUs while its input and KV cache may live on a different GPU, so tokens must be exchanged using the all-to-all communication primitive. This mismatch cannot be avoided without disconnecting the attention computation from the FFN computation, which would significantly degrade model quality (see Appendix~\ref{sec:gradient}).

Federation of Experts (FoE) addresses this bottleneck structurally by increasing the number of GPUs where a token's attention computation is performed. By doing this, fewer tokens need to be sent off-GPU for their selected experts, and more can be resolved locally \textit{without} all-to-all. FoE achieves this by replacing the single expert-parallel MoE block in each transformer layer with $H$ independent MoE clusters, which we call \emph{expert groups}. Each group owns $1/H$ of the KV heads, $1/H$ of the Q heads, $1/H$ of the experts, and $1/H$ of the GPUs. Routing is per-group: each token selects $k/H$ experts from each group rather than $k$ from the full pool, so every group contributes to every token at every layer. The $H$ groups synchronize once per layer through a cross-group mean of their post-attention residuals; dispatch and combine all-to-alls are strictly intra-group. Figure~\ref{fig:moe-vs-foe} contrasts this with MoE: one global all-to-all over $G$ GPUs is replaced by $H$ disjoint intra-group all-to-alls (each over $G/H$ GPUs) plus a much smaller cross-group all-reduce.

The central quantity for analysis is the local activation rate $\rho$, the fraction of expert selections resolved on the token's local GPUs (and which therefore require no all-to-all). Notation used throughout is summarized in Table~\ref{tab:notation} in the appendix. By splitting the KV cache across $H$ expert groups, FoE changes the structure of the expert-GPU assignment problem, increasing the minimum probability that a token's selected experts are co-located with its attention computation. While many prior works attempt to analyze historical expert routing patterns and speculatively compute a better expert-GPU assignment to thereby increase $\rho$, FoE guarantees a much higher local activation rate (LAR) by design.

In a standard MoE, the feasibility condition for full local resolution is $k \leq E/G$. For fine-grained MoE models with high $k$ deployed across large $G$, $E/G$ becomes small and this condition fails for most tokens, forcing them through the all-to-all for most of their $k$ selections. FoE relaxes the condition by $H\times$: the per-group analogue is $k/H \leq E/G$, structurally raising $\rho$ and reducing all-to-all volume in any deployment.

\subsection{Architecture Formulation}

We formalize FoE as a modification of a standard MoE layer, compared step-by-step in Table~\ref{tab:foe-vs-vanilla}. While a MoE keeps a single representation $x \in \mathbb{R}^{S \times d}$ at every layer boundary, FoE carries an $H$-fold representation $x \in \mathbb{R}^{H \times S \times d}$ between intermediate layers, one input per expert group, collapsing to $\mathbb{R}^{S \times d}$ only at the initial input embedding and after the final MoE layer's cross-group average. Keeping the expert group representations separate is what confines each group's computation to its own subset of experts and GPUs.

\begin{table}[h]
  \centering
  \small
  \setlength{\tabcolsep}{3pt}
  \renewcommand{\arraystretch}{1.15}
  \begin{tabular}{lll}
    \toprule
                     & Mixture of Experts                                                              & Federation of Experts                                                                                              \\
                     & $x \in \mathbb{R}^{S \times d}$                                                 & $x \in \mathbb{R}^{H \times S \times d}$                                                                           \\
    \midrule
    Attn.\ residual  & $r[i] = x[i] + \text{Attn}(x)[i]$                                               & $r[h, i] = x[h, i] + \text{Attn}_h(x[h, :])[i]$                                                                    \\
    Cross-group avg. & ---                                                                             & $r_{\text{full}}[i] = \tfrac{1}{H}\sum_{h} r[h, i]$                                                                \\
    Routing          & $\mathcal{I}[i],\, w[i] = \text{Router}(r[i])$                                  & $\mathcal{I}_h[i],\, w_h[i] = \text{Router}(r_{\text{full}}[i])$                                                   \\
    MoE update       & $x'[i] = r[i] + \sum\limits_{e \in \mathcal{I}[i]} w[i,e]\, \text{FFN}_e(r[i])$ & $x'[h, i] = r_{\text{full}}[i] + \sum\limits_{e \in \mathcal{I}_h[i]} w_h[i,e]\, \text{FFN}_e(r_{\text{full}}[i])$ \\
    \bottomrule
  \end{tabular}
  \caption{Per-layer compute for MoE and FoE on an intermediate layer ($1 < l < L$). All equations are per token $i \in [S]$. The first and last MoE layers in FoE differ only at the input/output boundaries (see text).}
  \label{tab:foe-vs-vanilla}
\end{table}

\textbf{Layer boundaries.} The first MoE layer ($l = 1$) receives the embedding $x^{(1)} \in \mathbb{R}^{S \times d}$ (no per-group dimension), runs a single shared attention to produce one residual $r$, and then applies the routing and MoE-update with $r$ in place of $r_{\text{full}}$ to produce $x^{(2)} \in \mathbb{R}^{H \times S \times d}$. The last MoE layer ($l = L$) is identical to the intermediate case but appends a final cross-group average $x^{(L+1)}[i] = \tfrac{1}{H}\sum_h x'[h, i]$ to project back to a single $\mathbb{R}^{S \times d}$ representation before the final linear token head.

\textbf{Routing.} Each expert group $h$ owns the contiguous block of experts $\mathcal{E}_h = \{h \cdot E/H,\, \ldots,\, (h{+}1) \cdot E/H - 1\}$. The router produces a single global score vector $g[i] \in \mathbb{R}^E$, and each group independently selects its top-$k/H$ experts from within its own expert group:
\[
  g[i] = \text{Router}(r_{\text{full}}[i]), \qquad
  \mathcal{I}_h[i],\, w_h[i] = \text{TopK}\bigl(g_{\mathcal{E}_h}[i],\, k/H\bigr) \quad \forall h \in [H],\, i \in [S].
\]

There is a \emph{single} router: one linear layer followed by one softmax over all $E$ experts, replicated identically on every GPU and evaluated on the shared representation $r_{\text{full}}[i]$. Only the top-$k/H$ \emph{selection} is partitioned per group, which is what enforces load balance; the scores themselves are global. Because every expert's score is normalized against all $E$ others, global softmax expert competition is retained exactly as in MoE.

\textbf{Convergence and expressivity.} A natural concern is that partitioning experts into groups may starve cross-expert information sharing and degrade convergence or expressivity relative to MoE. This is settled by direct proof, not just design intuition. Proposition~\ref{prop:jacobian} (Appendix~\ref{sec:gradient}) derives the \emph{per-expert inter-layer Jacobian}, the gradient of one expert's layer-$l$ output with respect to the previous layer's expert inputs. FoE replaces MoE's single block $J_e^{(l)}(\mathbf{I} + A_{ii}^{(l)})(M^{(l-1)} + R^{(l-1)})$ with the \emph{mean} of $H$ such blocks, one per group, and MoE is recovered exactly at $H = 1$. Three properties follow. The leading term is $\frac{1}{H}\sum_h \mathbf{I} = \mathbf{I}$, so the residual path that keeps deep MoE trainable survives intact. The block is a mean rather than a sum, so its magnitude is preserved and the gradient neither explodes nor vanishes as groups are added. Finally, every group enters symmetrically, so all $H$ attentions and all $k$ previously selected experts feed every current expert. FoE's gradient is therefore not merely similar to MoE's; it is the same formula averaged over groups. Ablations in the appendix confirm the design is necessary rather than incidental: removing the cross-group mean makes the groups independent models with zero cross-group gradient.

\subsection{Communication Analysis}

This section analyzes the communication volume and latency of a FoE architecture compared to a standard MoE architecture, decomposing traffic into intra-node and inter-node components. Under balanced routing, effective per-layer A2A volume is $2k(1-\rho) \cdot S \cdot d$. Mixture of Experts achieves $\rho_{\text{moe}} = 1/G$, while Federation of Experts achieves $\rho_{\text{foe}} = \min(H/G, 1)$, because each group only needs to resolve $k/H$ selections per token.

Both architectures synchronize once per layer, but not with the same primitives: each expert layer has dispatch and combine all-to-alls, and FoE adds one cross-group all-reduce. The two primitives differ in complexity. An \emph{all-to-all} requires each device to send a \emph{different} message to every other device, so nothing can be combined in transit and the volume each device ships grows in proportion to $k$. An \emph{all-reduce} does not: every participant needs the \emph{same} combined result, so partial results merge as they travel and each device sends a fixed amount regardless of $k$. FoE confines its all-to-all within a group and pays only the all-reduce between groups. Per layer, in units of $S \cdot d$ and measured as aggregate egress across all participating GPUs we get:
\[
  V_{\text{moe}}^{\text{A2A}} = 2k \cdot \frac{G-1}{G}, \quad V_{\text{foe}}^{\text{A2A}} = 2k \cdot \frac{\max(G-H,\,0)}{G}, \quad V_{\text{foe}}^{\text{AR}} = \frac{2(\min(G,H) - 1)}{\min(G,H)}.
\]

\paragraph{Single-node deployment ($N = 1$).} In almost all single-node settings $G \leq H$, since $n_{kv}$ is at least $8$ in modern production models and few single-node deployments exceed $8$ GPUs. Every expert a token selects is then already resident on the node, FoE's all-to-all term vanishes ($\rho = 1$), and only the cross-group all-reduce remains. This matters most for non-production deployments (edge boxes, workstations, single-node serving) where MoE's $(G - 1)/G$ factor already pushes it close to the naive cost.

\paragraph{Multi-node deployment ($1 < N \leq H$).} FoE expert groups are assigned to GPUs so that each spans at most one node, which keeps dispatch and combine intra-node regardless of $N$. When $N < H$, several groups share a node and each takes an equal subset of its GPUs. MoE instead runs a single expert-parallel group across all $N$ nodes, so a fraction $(N-1)/N$ of its all-to-all crosses node boundaries, while FoE sends only the all-reduce. Appendix~\ref{sec:comm-derivation} derives both volumes in full.

Raw volumes understate the actual cost difference because inter-node bandwidth is typically much lower than intra-node. Let $r = B_{\text{intra}} / B_{\text{inter}}$ (on H100-class hardware $900$ GB/s NVLink against $400$ Gb/s InfiniBand gives $r \approx 18$); weighting inter-node volume by $r$ gives total per-layer time. Table~\ref{tab:multi-node-comm} summarizes both raw aggregate volumes and bandwidth-weighted total time.

\begin{table}[h]
  \centering
  \small
  \setlength{\tabcolsep}{5pt}
  \renewcommand{\arraystretch}{1.15}
  \begin{tabular}{lccc}
    \toprule
    Deployment & Mixture of Experts & Federation of Experts & Reduction \\
    \midrule
    Single node ($N = 1$) & $2k \cdot \frac{G-1}{G}$ & $2 \cdot \frac{G-1}{G}$ & $k$ \\
    Multi-node ($N > 1$)  & $\frac{2k}{N} \cdot \frac{G-1}{G} \cdot \bigl[1 + r(N-1)\bigr]$ & $2k \cdot \frac{G-H}{G} + \frac{2(H-1)}{H} \cdot \frac{1 + r(N-1)}{N}$ & $\Theta(\min(k, r))$ \\
    \bottomrule
  \end{tabular}
  \caption{Per-layer communication time by deployment regime, in units of $S \cdot d$ with inter-node traffic weighted by $r$. The single-node row is not the multi-node formula at $N = 1$: the cross-group all-reduce then spans $G$ rather than $H$ participants.\label{tab:multi-node-comm}}
\end{table}

Thus in multi-node deployments, FoE relocates inter-node traffic: the $k$-proportional dispatch still happens, but within a group, on fewer GPUs and on a faster fabric, while the only traffic crossing the slow one is an all-reduce whose volume is independent of $k$. The total-time ratio has a clean asymptote, derived in Appendix~\ref{sec:comm-derivation}: for $k = H = 8$ and $r = 18$ it is $\approx 6.1\times$, within $\sim 5\%$ of the exact formula for $N \geq 2$.

\paragraph{Topology adaptability.}
FoE fixes only one quantity at training time, the group count $H = n_{kv}$, and that is a property of the \emph{model}, not the hardware. Nothing in the weights depends on the node count $N$ or the GPUs-per-node $G_n$. Groups are assigned to physical GPUs at deployment, subject only to spanning as few devices as possible. A single checkpoint therefore runs on any topology. Existing techniques like device-limited routing invert this: by capping the devices a token may reach, it bakes a node-count assumption into training-time routing. Evaluating the total-time ratio of Table~\ref{tab:multi-node-comm} for an $H = 8$ model on $4 \times 8$ and on $8 \times 4$ GPUs (both $G = 32$) gives $5.91\times$ and $6.19\times$: the same weights, two layouts, comparable benefit. Across $N \in \{1,2,4,8\}$ and $G_n \in \{2,4,8\}$ the predicted ratio stays within $5.79\times$--$8.00\times$, so FoE is never worse than MoE on any layout considered; Appendix~\ref{sec:topology} gives the full sweep. The gain does not require a hierarchical cluster: reducing the $k$-proportional all-to-all in favour of a $k$-independent all-reduce cuts volume even when every expert is equidistant. Hierarchy only compounds it, since MoE's flat all-to-all then pushes that larger volume onto the slowest link as well.

\section{Experiments}

\subsection{Experimental Setup}
\textbf{Hardware.} We evaluate performance across two deployment regimes: a single-node configuration utilizing an 8-GPU server equipped with NVIDIA H100 (SXM, HBM3) GPUs with 80GB of VRAM each, and a multi-node configuration consisting of two identical 8-GPU servers connected via InfiniBand. Both servers feature dual-socket Intel Xeon CPUs and 2TB of DDR5 memory.

\textbf{Inference Framework.} We implement both MoE and FoE within \textbf{FlexServe}, our inference engine based on FlexAttention \citep{Dong2024FlexAA}, which we will open source. Group Parallelism (GP) refers to FoE's cross-group all-reduce operations described in Section~\ref{sec:design}; FoE uses both GP and Expert Parallelism (EP), while MoE uses only EP. Running both architectures in one engine is deliberate: it attributes the measured difference to the architecture rather than to framework-level engineering. FlexServe integrates state-of-the-art inference techniques, including PagedAttention, chunked prefill/piggyback serving, data-parallelism, and pipeline parallelism \citep{kwon2023vllm, zheng2024sglang}. All measurements use the same pretrained 7B checkpoints; for the vLLM baseline we convert our MoE weights to vLLM's format so that the identical model is served and only the engine differs. Appendix~\ref{sec:model-spec} gives the full training configuration.

\begin{figure}[h]
    \centering
    \begin{subfigure}[b]{0.32\textwidth}
        \centering
        \includegraphics[width=\textwidth]{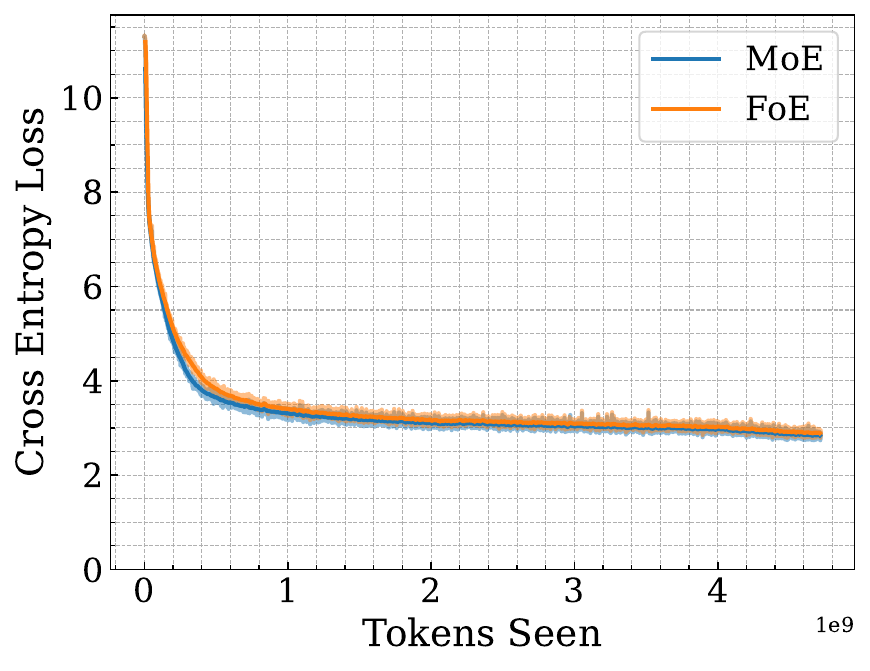}
        \caption{1B Model Training Loss}
        \label{fig:training-loss-1b}
    \end{subfigure}
    \hspace{0.05\textwidth}
    \begin{subfigure}[b]{0.32\textwidth}
        \centering
        \includegraphics[width=\textwidth]{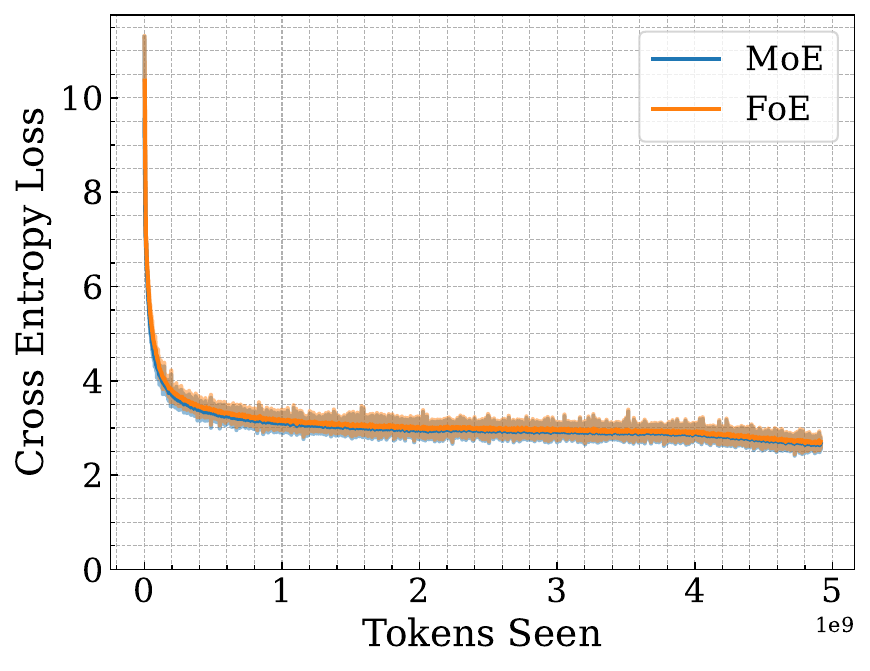}
        \caption{7B Model Training Loss}
        \label{fig:training-loss-7b}
    \end{subfigure}
    \caption{Cross-entropy (CE) loss curves at the end of pre-training. The FoE and MoE curves are almost indistinguishable at the end training. The 1B models are trained to Chinchilla-level; the 7B model is trained to $5\times$ its active parameter count \citep{chinchilla, olmoe}. The already small gap reduces with larger model size.}
    \label{fig:training-loss}
\end{figure}

\textbf{Model Pre-Training.} Because FoE changes the architecture, existing MoE checkpoints cannot be evaluated off-the-shelf. To empirically validate that restricted routing does not degrade convergence, we pretrain two OLMoE-like models from scratch alongside equivalently-sized MoE baselines \citep{olmoe}. FlexServe trains as well as serves, so a new architecture can be pretrained and then served for evaluation within one framework. Figure~\ref{fig:training-loss} visualizes the cross-entropy training loss for both configurations, confirming that the localized routing constraint of FoE maintains a similar convergence trajectory to MoE. We provide additional model and training details in Appendix~\ref{sec:appendix}.

\subsection{Evaluation Baselines and Metrics}

\textbf{Baselines.} We use two MoE baselines. The primary is expert parallelism (EP) with global all-to-all dispatch, implemented inside FlexServe. We also serve the \emph{identical} MoE checkpoint with vLLM \citep{kwon2023vllm}, a production system, to demonstrate performance against a heavily optimized engine with custom kernels. FoE could be implemented in vLLM, but this would require additional engineering effort which we leave to future work. Expert-placement and scheduling frameworks such as MoETuner \citep{moetuner}, Sem-MoE \citep{semmoe}, and EPLB-style rebalancers are \emph{orthogonal}: FoE still runs EP inside each group, so they apply within a group and compose on top of it. We therefore exclude them from latency comparisons and cross-compare on Local Activation Rate (LAR) instead which is a more direct measure of communication efficiency.

\textbf{Metrics.} Unlike many prior works which rely on purely synthetic evaluation, we model real-world request distributions with a Poisson arrival process on LongBench \citep{longbench}. We sweep the \emph{arrival scale} $1/\lambda$, the mean inter-arrival gap; a larger scale means lighter offered load. We separate prefill from decode using Time to First Token (TTFT, request arrival to first generated token) and Time per Output Token (TPOT, mean time per generated token during decode).

\subsection{End-to-End Inference Performance}
We measure TTFT and TPOT on the trained 7B models. Table~\ref{tab:inference-performance} reports mean, p50, and p99 for all three serving stacks. FoE substantially reduces TTFT by up to $\TTFTSpeedupSingle\times$ against the same-engine MoE baseline and by $\TTFTSpeedupVLLM\times$ against vLLM, and bounds outlier tails even better. This emphasizes FoE's inherent advantage in managing tail latencies by enforcing localized load-balancing guarantees without waiting on global stragglers (see section \ref{sec:lar_analysis}). At the lightest load the TPOT gap against the same-engine baseline narrows, since per-token cost is then dominated by fixed overheads rather than communication; the architectural advantage is a scaling effect and grows with offered load. Appendix~\ref{sec:prefill} isolates that scaling without queueing: FoE's prefill latency scales far more gracefully than MoE's, and the gap widens with batch size.

\begin{table}[h]
    \centering
    \small
    \begin{tabular}{llcccccc}
        \toprule
        & & \multicolumn{3}{c}{\textbf{TTFT (s)}} & \multicolumn{3}{c}{\textbf{TPOT (s)}} \\
        \cmidrule(lr){3-5} \cmidrule(lr){6-8}
        \textbf{Scale} & \textbf{Serving stack} & \textbf{Mean} & \textbf{p50} & \textbf{p99} & \textbf{Mean} & \textbf{p50} & \textbf{p99} \\
        \midrule
        0.25 & MoE (FlexServe) & 2.591 & 2.066 & 22.416 & 0.662 & 0.518 & 0.784 \\
             & MoE (vLLM)      & 3.425 & 2.623 & 38.952 & 0.537 & 0.525 & 0.981 \\
             & FoE (FlexServe) & 0.943 & 0.831 & 3.315 & 0.431 & 0.430 & 0.483 \\
             & \textit{Speedup} & \textit{3.63$\times$} & \textit{3.16$\times$} & \textit{11.75$\times$} & \textit{1.25$\times$} & \textit{1.22$\times$} & \textit{2.03$\times$} \\
        \midrule
        0.5 & MoE (FlexServe) & 1.687 & 1.624 & 3.287 & 0.535 & 0.560 & 0.721 \\
            & MoE (vLLM)      & 2.251 & 2.092 & 6.702 & 0.568 & 0.594 & 0.769 \\
            & FoE (FlexServe) & 0.620 & 0.606 & 1.058 & 0.371 & 0.372 & 0.390 \\
            & \textit{Speedup} & \textit{3.63$\times$} & \textit{3.45$\times$} & \textit{6.33$\times$} & \textit{1.53$\times$} & \textit{1.60$\times$} & \textit{1.97$\times$} \\
        \midrule
        0.75 & MoE (FlexServe) & 1.476 & 1.427 & 3.051 & 0.450 & 0.451 & 0.574 \\
             & MoE (vLLM)      & 1.872 & 1.772 & 4.236 & 0.487 & 0.494 & 0.639 \\
             & FoE (FlexServe) & 0.572 & 0.559 & 0.910 & 0.338 & 0.341 & 0.350 \\
             & \textit{Speedup} & \textit{3.27$\times$} & \textit{3.17$\times$} & \textit{4.65$\times$} & \textit{1.44$\times$} & \textit{1.45$\times$} & \textit{1.83$\times$} \\
        \bottomrule
    \end{tabular}
    \caption{Single-node inference performance across request arrival scales; a \emph{larger} scale means less frequent arrivals and lighter offered load. \textit{Speedup} is FoE over MoE (vLLM); ratios against MoE (FlexServe), i.e.\ the architecture-matched control in the same engine, are given in the text. TTFT: time to first token; TPOT: time per output token.}
    \label{tab:inference-performance}
\end{table}

\subsection{Scaling to Multi-Node Distributed Inference}
Modern deployments often exceed a single node, forcing all-to-all traffic onto the slower inter-node fabric. Table~\ref{tab:multi-node-performance} characterizes inference performance on a multi-node configuration consisting of two identical 8-GPU servers running the trained 7B models across the same three serving stacks.

FoE scales gracefully to larger deployments and, as its design predicts, gains \textit{more} from moving single-node to multi-node than the MoE baseline does. At an arrival scale of $0.5$, FoE's TTFT speedup over the same-engine baseline rises from $\TTFTSpeedupSingleAtHalf\times$ single-node to $\TTFTSpeedupMultiAtHalf\times$ multi-node. This is because MoE must dispatch tokens across the whole combined topology and pays the interconnect penalty, whereas FoE confines dispatch and combine strictly within localized expert groups instead of on the slow inter-node fabric. Tails behave the same way: at that scale FoE's p99 TTFT stays at $\TailTTFTFoEMulti$s against MoE's inflated $\TailTTFTMoEMulti$s.

\begin{table}[t]
    \centering
    \small
    \begin{tabular}{llcccccc}
        \toprule
        & & \multicolumn{3}{c}{\textbf{TTFT (s)}} & \multicolumn{3}{c}{\textbf{TPOT (s)}} \\
        \cmidrule(lr){3-5} \cmidrule(lr){6-8}
        \textbf{Scale} & \textbf{Serving stack} & \textbf{Mean} & \textbf{p50} & \textbf{p99} & \textbf{Mean} & \textbf{p50} & \textbf{p99} \\
        \midrule
        0.25 & MoE (FlexServe) & 2.156 & 2.049 & 4.060 & 0.515 & 0.506 & 0.842 \\
             & MoE (vLLM)      & 2.566 & 2.450 & 5.058 & 0.503 & 0.484 & 0.989 \\
             & FoE (FlexServe) & 0.701 & 0.689 & 1.110 & 0.413 & 0.414 & 0.445 \\
             & \textit{Speedup} & \textit{3.66$\times$} & \textit{3.56$\times$} & \textit{4.56$\times$} & \textit{1.22$\times$} & \textit{1.17$\times$} & \textit{2.22$\times$} \\
        \midrule
        0.5 & MoE (FlexServe) & 1.702 & 1.610 & 3.375 & 0.528 & 0.540 & 0.708 \\
            & MoE (vLLM)      & 2.086 & 1.827 & 7.184 & 0.515 & 0.530 & 0.764 \\
            & FoE (FlexServe) & 0.587 & 0.584 & 0.872 & 0.369 & 0.367 & 0.385 \\
            & \textit{Speedup} & \textit{3.55$\times$} & \textit{3.13$\times$} & \textit{8.24$\times$} & \textit{1.40$\times$} & \textit{1.44$\times$} & \textit{1.98$\times$} \\
        \bottomrule
    \end{tabular}
    \caption{Multi-node inference performance on two identical 8-GPU servers connected via InfiniBand. Both TTFT and its tail scale efficiently because localized routing shields dispatch traffic from the slower inter-node fabric. Scale and \textit{Speedup} are defined as in Table~\ref{tab:inference-performance}.}
    \label{tab:multi-node-performance}
\end{table}

\subsection{Local Activation Rate and GPU Load Imbalance}
\label{sec:lar_analysis}

A critical factor driving FoE's communication efficiency is its ability to maximize the Local Activation Rate (LAR) while simultaneously maintaining perfect load balancing.

\begin{figure}[h]
    \centering
    \begin{subfigure}[b]{0.45\linewidth}
        \centering
        \begin{tikzpicture}
            \begin{axis}[
                ybar,
                symbolic x coords={EP, MoETuner, Sem-MoE, FoE (Ours)},
                xtick=data,
                ylabel={Local Activation Rate},
                ymin=0, ymax=1.1,
                bar width=10pt,
                width=0.9\linewidth,
                height=3.5cm,
                enlarge x limits=0.2,
                x tick label style={font=\scriptsize,rotate=45,anchor=east},
                y tick label style={font=\scriptsize},
                ylabel style={font=\scriptsize}
            ]
                \addplot[fill=blue!30] coordinates {
                    (EP, 0.125)
                    (MoETuner, \LARMoETuner)
                    (Sem-MoE, \LARSemMoE)
                    (FoE (Ours), 1.00)
                };
            \end{axis}
        \end{tikzpicture}
        \caption{Local Activation Rate (LAR)}
        \label{fig:lar_comparison}
    \end{subfigure}
    \hfill
    \begin{subfigure}[b]{0.45\linewidth}
        \centering
        \begin{tikzpicture}
            \begin{axis}[
                ybar,
                symbolic x coords={MoETuner, Sem-MoE, FoE (Ours)},
                xtick=data,
                ylabel={Max/Median Discrepancy},
                ymin=0.5,
                bar width=10pt,
                width=0.9\linewidth,
                height=3.5cm,
                enlarge x limits=0.2,
                x tick label style={font=\scriptsize,rotate=45,anchor=east},
                y tick label style={font=\scriptsize},
                ylabel style={font=\scriptsize}
            ]
                \addplot[fill=red!30] coordinates {
                    (MoETuner, 1.67)
                    (Sem-MoE, 1.40)
                    (FoE (Ours), 1.00)
                };
            \end{axis}
        \end{tikzpicture}
        \caption{GPU Load Imbalance Rate}
        \label{fig:imbalance_comparison}
    \end{subfigure}
    \caption{Side-by-side assessment of structural efficiency for a single node with 8 GPUs. (a) FoE structurally guarantees all tokens are processed locally, achieving a perfect LAR of 1.0. (b) Comparison of GPU load discrepancy (max/median token assignments), with values for MoETuner and Sem-MoE sourced from their papers, highlighting FoE's perfect load balancing.}
    \label{fig:efficiency_comparison}
\end{figure}
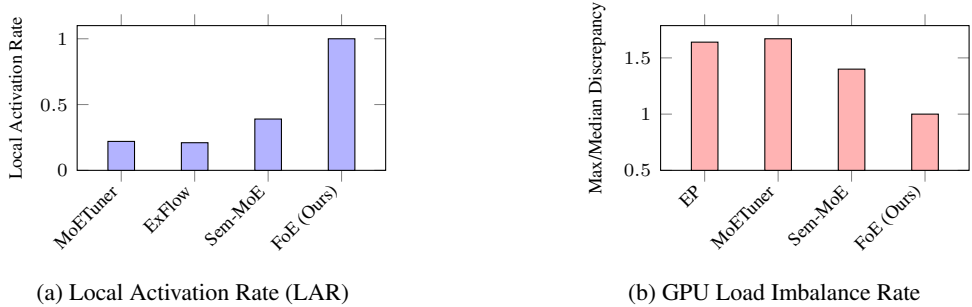

In Figure~\ref{fig:efficiency_comparison}(a), we compare the LAR of our FoE architecture against prior state-of-the-art predictive and balancing frameworks (Sem-MoE, MoETuner) at an Expert Parallel (EP) size of 8. Because FoE structurally confines routing within isolated expert groups (in our single-node configuration of 8 GPUs, each group is self-contained), it inherently achieves an LAR of 1.0 (100\%). In contrast, heuristic-based proxy methods suffer from structural degradation as EP size scales; at EP=8, Sem-MoE achieves an LAR of \LARSemMoE, and MoETuner drops to \LARMoETuner.

Simultaneously, traditional MoE systems relying on global all-to-all routing frequently suffer from acute token processing imbalances across GPUs. We define this GPU load discrepancy as the ratio between the maximum and median number of tokens routed to a GPU during inference. Our execution traces (Appendix~\ref{sec:comm_traces}) show that standard MoE suffers an average max/median GPU load discrepancy of $\LoadDiscrepancyMoE\times$ in a given forward pass. Because standard expert parallelism synchronizes at a global all-to-all, every GPU must wait for the slowest straggler. 

Because FoE routes every token to every expert group, all groups receive the same number of tokens. In a single-node setting every GPU therefore does too, giving a structural load-balance rate of $1.0$. Figure~\ref{fig:imbalance_comparison} contrasts this with optimized placement baselines, whose values are taken from their respective publications. The serving gains reported above thus come from both effects together: traffic that never leaves the group, and no straggler barrier, since every group receives the same number of tokens.

This advantage is not confined to the converged model. Over the whole pretraining run, FoE holds a lower KL divergence from uniform than its size-matched MoE baseline at every checkpoint, with the load-balancing loss coefficient left at the baseline setting (Appendix~\ref{sec:balance}). Balance in FoE is therefore structural, not recovered by tuning an auxiliary loss.

\subsection{Generation Quality}
Table~\ref{tab:generation-quality} confirms that structurally separating experts into subgroups does not significantly degrade model quality. We compare models pre-trained from scratch under identical settings. FoE trails its size-matched MoE baseline closely on most tasks and leads on OpenBookQA at 7B, a spread within the range of training seed variance at this scale. FoE's communication savings therefore do not come at a material cost in model quality.

\begin{table}[h]
    \centering
    \small
    \setlength{\tabcolsep}{4pt}
    \begin{tabular}{lccccccc}
        \toprule
        \textbf{Model Variant} & \textbf{Lambada} & \textbf{SciQ} & \textbf{HellaSwag}$^{\dagger}$ & \textbf{ARC-Easy} & \textbf{PIQA}$^{\dagger}$ & \textbf{Social IQa} & \textbf{OpenBookQA}$^{\dagger}$ \\
        \midrule
        MoE (7B)  & 33.5\% & 82.6\% & 43.8\% & 60.6\% & 68.2\% & 39.0\% & 34.2\% \\
        FoE (7B)  & 32.5\% & 80.8\% & 42.1\% & 60.3\% & 67.2\% & 37.6\% & 34.8\% \\
        \midrule
        MoE (1B)  & 29.0\% & 76.5\% & 36.1\% & 55.6\% & 65.1\% & 39.0\% & 33.6\% \\
        FoE (1B)  & 27.2\% & 75.4\% & 34.9\% & 54.1\% & 63.8\% & 37.2\% & 32.4\% \\
        \bottomrule
    \end{tabular}
    \caption{Zero-shot accuracy across evaluation benchmarks. $^{\dagger}$ marks length-normalised accuracy; the rest are plain accuracy. Benchmarks were selected for signal at this scale: each is several standard errors above its naive baseline.}
    \label{tab:generation-quality}
\end{table}

\section{Limitations}

While FoE significantly reduces communication overhead, its benefits only apply to distributed deployments: on a single GPU there is no all-to-all to eliminate, and the cross-group all-reduce becomes overhead. In distributed deployments, the placement of expert groups across nodes determines how much traffic stays intra-node. Each group is placed on as few nodes as possible, which is always achievable, and Section~\ref{sec:design} shows a single checkpoint retains its advantage across layouts. Genuinely heterogeneous clusters, whose nodes differ in size or interconnect bandwidth, would need placement-aware assignment that we do not study here. The only architectural coupling FoE introduces is a divisibility requirement: the group count is set to $n_{kv}$, and the expert count, the top-$k$, and the query heads must each be divisible by it. Grouped-query attention alone forces the last of these, and our survey of production MoEs (Appendix~\ref{sec:divisibility}) finds the conditions met in almost every model we examined. FoE adds no training hyperparameters, trains stably under settings identical to its MoE baseline, and is more load-balanced than standard MoE by construction (Section~\ref{sec:lar_analysis}).

A second limitation is that FoE cannot accelerate an existing MoE checkpoint. Grouping changes how experts and KV heads are partitioned, so a model has to be trained in the FoE configuration, as ours were. We regard knowledge distillation as the natural route to reusing already-trained models: FoE is structurally close to MoE, so expert-group assignments can be chosen from the historical activation patterns of an MoE teacher and then training continues in the FoE configuration. Distillation has already bridged far larger architectural gaps, including attention models into state-space models \citep{junxiongdaniele2024mambainllama} and dense models into MoE \citep{komatsuzaki2023sparse}, which makes MoE to FoE a comparatively small step. Doing it properly requires its own manuscript, and we leave it to future work.

Finally, our empirical evaluation covers two model sizes (1B and 7B) and two hardware configurations (a single $8\times$H100 node and a two-node InfiniBand cluster). Training at larger scale was beyond our budget.

\section{Conclusion}

We present Federation of Experts (FoE), a novel architectural reformulation of Mixture-of-Experts that partitions each transformer layer into independent expert groups, replacing the global all-to-all of expert parallelism with several smaller intra-group all-to-alls plus a much cheaper cross-group all-reduce. This guarantees that more tokens resolve locally, substantially cutting communication overhead. Empirically, FoE delivers up to a $\FwdSpeedupFlex\times$ end-to-end prefill speedup, $\TTFTSpeedupBest\times$ lower TTFT, and $\TPOTSpeedupBest\times$ lower TPOT, with cross-entropy loss and zero-shot reasoning on par with the MoE baseline. The benefit compounds on multi-node deployments, where the intra-group all-to-alls stay on the fast intra-node fabric while only the cheaper all-reduce crosses the inter-node network. These gains are orthogonal to existing placement, scheduling, and prefetching optimizations, all of which still apply on top of FoE.

\clearpage

\section*{AI Use Statement}
During the preparation of this work, the authors used AI tools for the purpose of proofreading, editing and latex formatting. The text was written by the authors, and AI was only used to assist in the drafting of the text. Claude Code was also used to assist in software development, identifying issues, and finding background literature. The authors take full responsibility for the validity and correctness of all generated content. No AI tools were used to generate core ideas, obtain experimental results, or run experiments.

\section*{Ethics Statement}
No ethical issues were identified in the preparation of this work. The authors have no conflicts of interest to disclose.

\section*{Reproducibility Statement}
We will release the code and checkpoints used in this work to facilitate reproducibility. All experiments were run on publicly available datasets.

\clearpage

\bibliographystyle{iclr2027_conference}
\bibliography{references}

\appendix

\raggedbottom

\newpage
\section{Appendix}
\label{sec:appendix}

\subsection{Notation}

Table~\ref{tab:notation} collects the notation used throughout the paper, in particular in the design and communication analysis of Section~\ref{sec:design}.

\begin{table}[H]
  \centering
  \small
  \begin{tabular}{ll}
    \toprule
    Symbol                                  & Meaning                                                                                     \\
    \midrule
    $G$                                     & Total GPUs in the deployment                                                                \\
    $N$, $G_n = G/N$                        & Number of nodes, GPUs per node                                                              \\
    $H = n_{kv}$                            & Expert groups in FoE (set equal to the model's KV-head count)                               \\
    $L$, $E$, $k$                           & MoE layers, total routed experts, top-$k$ per token ($k$ a multiple of $H$)                 \\
    $S$, $d$                                & Tokens per batch, hidden dimension                                                          \\
    \midrule
    $i$, $h$, $l$, $e$                      & Token, group, layer, expert indices                                                         \\
    $x[i]$ / $x[h, i]$                      & Hidden state at a layer boundary (MoE / per-group in FoE)                                   \\
    $r[i]$ / $r[h, i]$                      & Post-attention residual (MoE / per-group in FoE)                                            \\
    $r_{\text{full}}[i]$                    & Cross-group averaged residual: $\tfrac{1}{H}\sum_h r[h, i]$                                 \\
    $g[i] \in \mathbb{R}^E$                 & Router scores over all $E$ experts                                                          \\
    $\mathcal{E}_h$                         & Expert index block owned by group $h$: $\{h \cdot E/H,\, \ldots,\, (h{+}1) \cdot E/H - 1\}$ \\
    $g_{\mathcal{E}_h}[i] \in \mathbb{R}^{E/H}$ & Sub-vector of $g[i]$ restricted to the expert block $\mathcal{E}_h$                     \\
    $\text{TopK}(v, m)$                     & Indices and values of the $m$ largest entries of vector $v$                                 \\
    $\mathcal{I}_h[i]$, $w_h[i]$            & Top-$k/H$ expert indices and routing weights for group $h$, token $i$                       \\
    \midrule
    $\rho$                                  & Local activation rate (fraction of expert selections resolved on-GPU)                       \\
    $V^{\text{A2A}}$, $V^{\text{AR}}$       & Per-layer all-to-all and all-reduce communication volumes (in units of $S \cdot d$)         \\
    $B_{\text{intra}}$, $B_{\text{inter}}$  & Intra-node (NVLink/NVSwitch/PCIe) and inter-node (InfiniBand/Ethernet) bandwidth            \\
    $r = B_{\text{intra}}/B_{\text{inter}}$ & Intra/inter-node bandwidth ratio ($r \approx 15$--$20$ on H100-class hardware)              \\
    $T$                                     & Bandwidth-weighted per-layer communication time                                             \\
    \bottomrule
  \end{tabular}
  \caption{Notation used throughout the Design section.\label{tab:notation}}
\end{table}

\subsection{Model Specification}
\label{sec:model-spec}

Table~\ref{tab:training-config} lists the architectural dimensions and training hyperparameters used for the small ($\sim$1B) and full ($\sim$7B) pretraining runs reported in the experiments section. Both models are OLMoE-style transformer MoEs pretrained from scratch on FineWeb-Edu using the OLMoE-1B-7B-0924 tokenizer, with $n_{kv} = 8$ KV heads (matching $H = 8$ expert groups in FoE). During inference, FoE uses a negligible amount of additional GPU memory, due to the increased activation memory from independent residual streams from each expert group.

\begin{table}[t]
    \centering
    \small
    \begin{tabular}{lll}
        \toprule
        Variant                          & Small                                          & Full               \\
        \midrule
        \multicolumn{3}{l}{\textit{Architecture}}                                                                \\
        Number of Parameters               & 1.02B                                          & 6.85B              \\
        Hidden dimension ($d$)             & 768                                            & 2{,}048            \\
        Number of layers ($L$)             & 16                                             & 16                 \\
        Max sequence length                & 32{,}768                                       & 32{,}768           \\
        Training sequence length           & 2{,}048                                        & 2{,}048            \\
        Vocabulary                         & \multicolumn{2}{l}{OLMoE-1B-7B-0924 tokenizer}                      \\
        \midrule
        \multicolumn{3}{l}{\textit{Attention}}                                                                   \\
        Query heads                        & 8                                              & 16                 \\
        KV heads ($n_{kv}$)                & 8                                              & 8                  \\
        Head dimension                     & 96                                             & 128                \\
        RoPE base ($\theta$)               & 500{,}000                                      & 500{,}000          \\
        RoPE scaling factor                & $16\times$                                     & $16\times$         \\
        \midrule
        \multicolumn{3}{l}{\textit{MoE FFN}}                                                                     \\
        Total routed experts ($E$)         & 64                                             & 64                 \\
        Top-$k$ experts                    & 8                                              & 8                  \\
        Expert FFN dim multiplier          & $1/2$                                          & $1/2$              \\
        Expert groups (FoE only, $H$)      & 8                                              & 8                  \\
        \midrule
        \multicolumn{3}{l}{\textit{Training}}                                                                    \\
        Dataset                            & \multicolumn{2}{l}{FineWeb-Edu (train split)}                       \\
        Local batch size (per DP instance) & 16                                             & 6                  \\
        Data-parallel shard size           & 8                                              & 8                  \\
        Warmup steps                       & 1{,}000                                        & 1{,}000            \\
        Total steps                        & 18{,}000                                       & 50{,}000           \\
        Decay-start step                   & 15{,}000                                       & 40{,}000           \\
        Gradient-norm clip                 & 1.0                                            & 1.0                \\
        Total tokens trained on            & 4.7B                                           & 4.9B               \\
        \midrule
        \multicolumn{3}{l}{\textit{Optimizer (AdamW)}}                                                           \\
        Learning rate                      & $6 \times 10^{-4}$                             & $6 \times 10^{-4}$ \\
        Weight decay                       & 0.1                                            & 0.1                \\
        $(\beta_1, \beta_2)$               & $(0.9, 0.95)$                                  & $(0.9, 0.95)$      \\
        \midrule
        \multicolumn{3}{l}{\textit{Precision}}                                                                   \\
        Parameter dtype                    & bfloat16                                       & bfloat16           \\
        \bottomrule
    \end{tabular}
    \caption{Training configuration for the small ($\sim$1B) and full ($\sim$7B) FoE and size-matched MoE pretraining runs. Architectural rows that differ between architectures (e.g., expert groups) are marked accordingly; all other settings are shared between the FoE model and its MoE baseline.\label{tab:training-config}}
\end{table}

\subsection{Communication Volume Derivation}
\label{sec:comm-derivation}

This section derives the per-layer communication volumes of Tables~\ref{tab:single-node-comm} and~\ref{tab:comm-derivation-table} and states the deployment assumptions behind them.

Each experts layer has dispatch and combine all-to-alls (identical in volume), and for FoE, one additional cross-group all-reduce. The all-reduce runs across $\min(G, H)$ participants --- physically $H$-way when $G \geq H$ (each group occupies its own GPU or set of GPUs), and $G$-way when $G < H$ (multiple groups sharing a GPU combine partials locally before the network hop). Substituting $\rho_{\text{moe}} = 1/G$ and $\rho_{\text{foe}} = \min(H/G, 1)$ into the effective A2A volume $2k(1-\rho)$ gives:
\[
  V_{\text{moe}}^{\text{A2A}} = 2k \cdot \frac{G-1}{G}, \quad V_{\text{foe}}^{\text{A2A}} = 2k \cdot \frac{\max(G-H,\,0)}{G}, \quad V_{\text{foe}}^{\text{AR}} = \frac{2(\min(G,H) - 1)}{\min(G,H)}.
\]
We report communication volumes as aggregate egress across all participating GPUs in units of $S \cdot d$.

\paragraph{Single-node deployment ($N=1$).}
In almost all single-node settings $G \leq H$ because $n_{kv}$ is typically $8$ or more in modern production models and almost no single-node deployments have more than $8$ GPUs. We therefore restrict attention to the $G \leq H$ regime for single-node, under which the FoE A2A term vanishes ($\rho = 1$) and only the cross-group all-reduce (of volume $2(G-1)/G$) remains.

Per-layer intra-node volume is $2k \cdot \frac{G-1}{G}$ for MoE and $2 \cdot \frac{G-1}{G}$ for FoE (in units of $S \cdot d$), so the ratio collapses to $V_{\text{moe}}/V_{\text{foe}} = k$: FoE eliminates the $k$-proportional A2A term, and communication reduction on a single node is exactly $k\times$ for \emph{any} $G \leq H$, whether the node matches the model exactly ($G = H$) or is under-provisioned relative to the model ($G < H$). This is particularly valuable for non-production scale deployments (edge boxes, workstations, single-node inference) where MoE's $(G - 1)/G$ factor pushes it close to the full $2k$ naive cost.

\begin{table}[H]
  \centering
  \small
  \setlength{\tabcolsep}{5pt}
  \renewcommand{\arraystretch}{1.15}
  \begin{tabular}{lccc}
    \toprule
               & Mixture of Experts & Federation of Experts & Reduction ($\text{MoE}/\text{FoE}$) \\
    \midrule
    Intra-node & $2k \cdot \frac{G-1}{G}$ & $2 \cdot \frac{G-1}{G}$ & $k$ \\
    \bottomrule
  \end{tabular}
  \caption{Per-layer communication volume (units of $S \cdot d$) for a single-node deployment ($N = 1$, $G \leq H$). There is no inter-node term, so total time equals the intra-node volume. FoE eliminates the $k$-proportional all-to-all entirely, leaving only the cross-group all-reduce over $\min(G,H) = G$ participants, so the reduction is exactly $k\times$ for any $G \leq H$. Laid out as Table~\ref{tab:comm-derivation-table} for comparison.\label{tab:single-node-comm}}
\end{table}

\paragraph{Multi-node deployment ($1 < N \leq H$).}
For FoE, experts are assigned to GPUs such that each expert group spans at most one node. This way dispatch and combine all-to-alls are intra-node regardless of $N$. When $N < H$, multiple groups share a node (each group gets an equal subset of the GPUs). MoE, in contrast, runs a single EP group across all $N$ nodes, so under balanced routing a fraction $(N-1)/N$ of its A2A crosses node boundaries. The same fraction applies to the FoE all-reduce when the ring is ordered so that same-node groups are consecutive. Multi-node deployments have $G \geq H$, so the FoE all-reduce term takes the $2(H-1)/H$ form, bounding inter-node volume above by $\approx 2$ regardless of $k$; MoE's inter-node volume, by contrast, grows linearly in $k$.

\begin{table}[H]
  \centering
  \small
  \setlength{\tabcolsep}{5pt}
  \renewcommand{\arraystretch}{1.15}
  \begin{tabular}{lccc}
    \toprule
               & Mixture of Experts & Federation of Experts & Reduction ($\text{MoE}/\text{FoE}$) \\
    \midrule
    Intra-node & $\frac{2k}{N} \cdot \frac{G-1}{G}$ & $2k \cdot \frac{G-H}{G} + \frac{1}{N} \cdot \frac{2(H-1)}{H}$ & $\Theta(1/N)$ \\
    Inter-node & $\frac{2k(N-1)}{N} \cdot \frac{G-1}{G}$ & $\frac{N-1}{N} \cdot \frac{2(H-1)}{H}$ & $\Theta(k)$ \\
    Total time & $\frac{2k}{N} \cdot \frac{G-1}{G} \cdot \bigl[1 + r(N-1)\bigr]$ & $2k \cdot \frac{G-H}{G} + \frac{2(H-1)}{H} \cdot \frac{1 + r(N-1)}{N}$ & $\Theta(\min(k, r))$ \\
    \bottomrule
  \end{tabular}
  \caption{Per-layer communication cost in a multi-node deployment ($N \leq H$), in units of $S \cdot d$, reproduced from Table~\ref{tab:multi-node-comm} so the derivation is self-contained. The first two rows give raw aggregate volumes; the last row weights inter-node by $r = B_{\text{intra}}/B_{\text{inter}}$ ($\approx 18$ on H100-class hardware) so it predicts wall-clock latency. The last column gives the asymptotic MoE/FoE ratio as $N$ grows: FoE cuts inter-node volume by $\Theta(k)$ and instead carries more intra-node traffic, so the bandwidth-weighted total improves by $\Theta(\min(k, r))$ and needs $r > 1$ to pay off.\label{tab:comm-derivation-table}}
\end{table}

The total-time ratio collapses to a clean asymptote:
\begin{align*}
  \frac{T_{\text{moe}}}{T_{\text{foe}}}
   & = \frac{\frac{2k}{N} \cdot \frac{G-1}{G} \cdot \bigl[1 + r(N-1)\bigr]}{2k \cdot \frac{G-H}{G} + \frac{2(H-1)}{H} \cdot \frac{1 + r(N-1)}{N}}
  \longrightarrow \frac{kHr}{kH + r(H-1)} \quad (N \to \infty,\ G = G_n N).
\end{align*}

For $k = 8$, $H = 8$, $r = 18$ the asymptote is $\approx 6.1\times$, which is within $\sim 5\%$ of the actual formula for $N \geq 2$.

\subsection{Detailed Communication Traces}
\label{sec:comm_traces}

\begin{figure}[H]
    \centering
    \includegraphics[width=\linewidth]{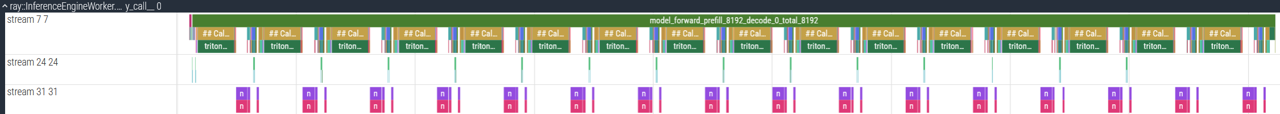}
    \caption{Detailed communication trace (flamegraph) for a single layer of a standard MoE baseline. Stream 7 is compute and stream 31 is communication.}
    \label{fig:moe_trace}
\end{figure}

\begin{figure}[H]
    \centering
    \includegraphics[width=\linewidth]{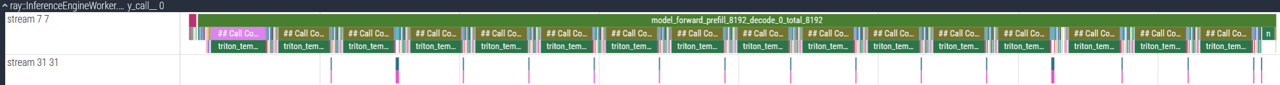}
    \caption{Detailed communication trace (flamegraph) for a single layer of the proposed FoE architecture. Stream 7 is compute and stream 31 is communication. Note the reduced communication overhead compared to the standard MoE trace in Figure~\ref{fig:moe_trace}.}
    \label{fig:foe_trace}
\end{figure}

To demonstrate the detailed communication patterns during training, we profile communication for a single layer of an 8{,}192-token batch run on an 8-GPU single node. Figure~\ref{fig:moe_trace} shows the trace for the standard MoE baseline, whereas Figure~\ref{fig:foe_trace} illustrates the trace for our FoE architecture. We can observe the significant reduction in synchronization overhead between the two designs. In particular, on our $8\times$H100 setup we measured a $\CommReduction\times$ reduction in communication time (stream 31).

\subsection{End-to-End Prefill Latency}
\label{sec:prefill}
\begin{figure}[H]
    \centering
    \includegraphics[width=\linewidth]{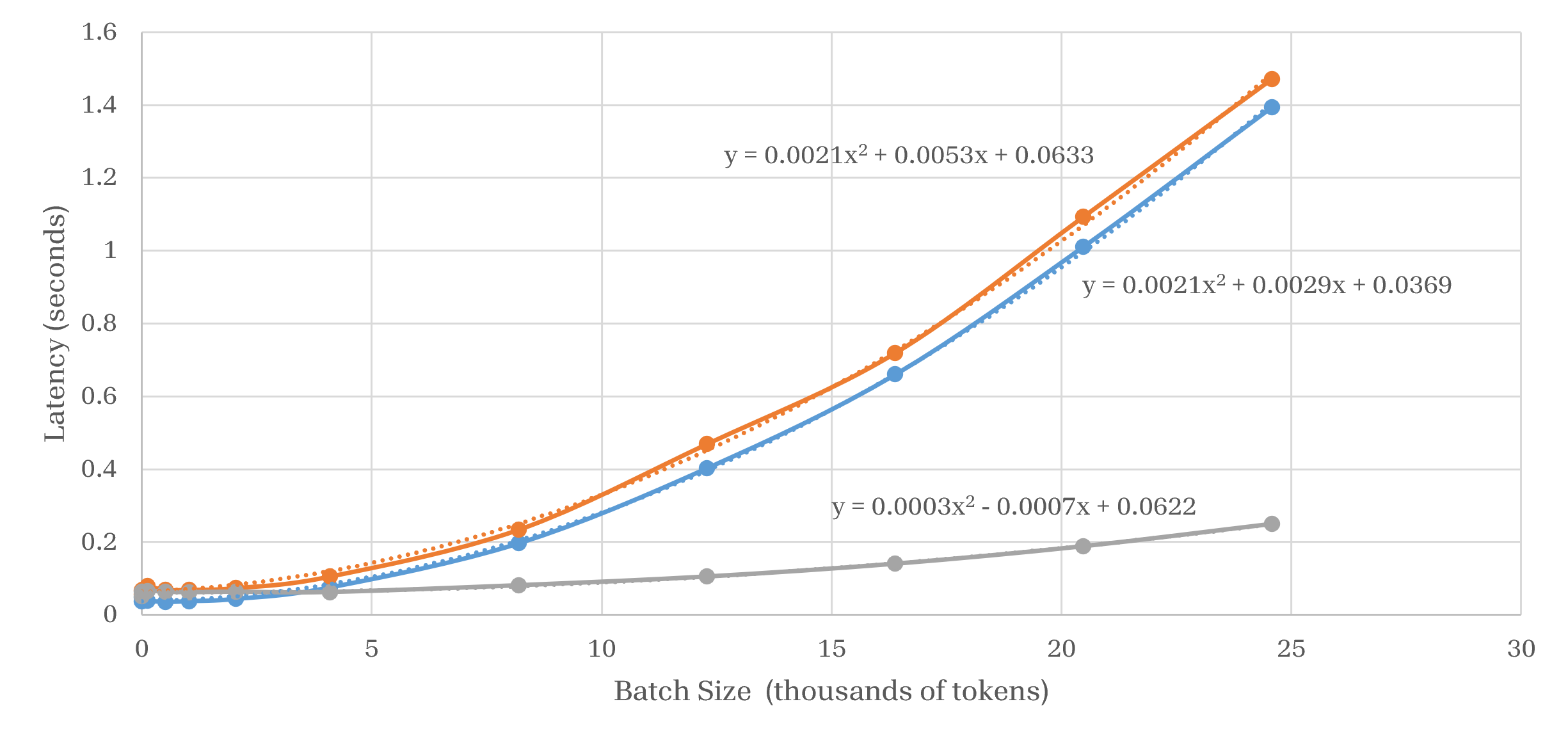}
    \caption{End-to-end prefill latency against batch size for the three serving stacks, with quadratic fits. Both MoE stacks share a quadratic coefficient of $\FitCoefMoE$, so FlexServe is a competitive baseline and the comparison isolates the architecture rather than the engine. FoE's coefficient is $\FitCoefFoE$, roughly seven times smaller, so its latency scales far more gracefully than MoE's and the gap widens with batch size.\label{fig:latency-sweep}}
\end{figure}

Table~\ref{tab:forward-latency} and Figure~\ref{fig:latency-sweep} report end-to-end prefill latency as the synthetic batch size increases, for FoE against an architecture-matched MoE in the same engine and against vLLM. In standard MoE parallelism, synchronization requires global dispatch across the network topology. As batch sizes expand beyond 2048, prefill latency escalates substantially for both MoE stacks, bottlenecked by the all-to-all communication overhead. In contrast, FoE confines routing and load balancing within localized expert groups, holding its quadratic growth coefficient to $\FitCoefFoE$ against $\FitCoefMoE$ for both MoE stacks, roughly seven times smaller, and yielding a $\FwdSpeedupFlex\times$ speedup over MoE (FlexServe) and $\FwdSpeedupVLLM\times$ over vLLM at a batch size of \FwdBatchMax. Below 4096 tokens, both FlexServe stacks trail vLLM by a roughly constant factor. That gap reflects vLLM's custom fused kernels for MoE compute, routing, and KV-cache operations rather than any architectural difference: it is constant in batch size, whereas FoE's advantage grows with it.

\begin{table}[H]
    \centering
    \small
    \begin{tabular}{rcccc}
        \toprule
        & \multicolumn{3}{c}{\textbf{Mean latency (s)}} & \\
        \cmidrule(lr){2-4}
        \textbf{Batch size} & \textbf{MoE (FlexServe)} & \textbf{MoE (vLLM)} & \textbf{FoE (FlexServe)} & \textbf{Speedup} \\
        \midrule
        1        & 0.058 & 0.047 & 0.051 & $1.13\times$ \\
        2        & 0.068 & 0.038 & 0.064 & $1.05\times$ \\
        8        & 0.070 & 0.042 & 0.065 & $1.08\times$ \\
        128      & 0.079 & 0.040 & 0.067 & $1.19\times$ \\
        512      & 0.069 & 0.036 & 0.064 & $1.08\times$ \\
        1{,}024  & 0.069 & 0.038 & 0.062 & $1.12\times$ \\
        2{,}048  & 0.074 & 0.045 & 0.063 & $1.17\times$ \\
        4{,}096  & 0.106 & 0.076 & 0.063 & $1.69\times$ \\
        8{,}192  & 0.234 & 0.197 & 0.081 & $2.88\times$ \\
        12{,}288 & 0.469 & 0.403 & 0.105 & $4.46\times$ \\
        16{,}384 & 0.719 & 0.661 & 0.141 & $5.12\times$ \\
        20{,}480 & 1.093 & 1.012 & 0.189 & $5.79\times$ \\
        24{,}576 & 1.471 & 1.395 & 0.250 & $5.88\times$ \\
        \bottomrule
    \end{tabular}
    \caption{End-to-end prefill latency (seconds, mean of three runs) for a single synthetic batch, measured across batch sizes. Each measurement covers one prefill step through the full serving stack, from request input to KV-cache write, and excludes queueing. \emph{Speedup} is MoE (FlexServe) over FoE (FlexServe), the architecture-matched comparison within one engine; this is a single prefill step rather than an end-to-end serving measurement, so the vLLM ratio is reported in the text instead.}
    \label{tab:forward-latency}
\end{table}

\subsection{Gradient Analysis: Convergence and Expressivity}
\label{sec:gradient}

A natural concern with partitioning experts into $H$ groups is that it might starve cross-expert information sharing, and so harm convergence or expressivity. This section shows it does not, by computing the \emph{per-expert inter-layer Jacobian}: the gradient of one expert's layer-$l$ output with respect to the shared representation the previous layer's experts take as input. We differentiate through the router as well, so the result is the full gradient rather than a frozen-routing approximation.

\paragraph{Setup.}
Fix a token $i$. Expert $e$'s layer-$l$ output is $o_e^{(l)}[i] = \text{FFN}_e^{(l)}(u_e^{(l)}[i])$, with input $u_e^{(l)}[i] = r^{(l)}[i]$ for MoE and $u_e^{(l)}[i] = r_{\text{full}}^{(l)}[i] = \frac{1}{H}\sum_h r^{(l)}[h,i]$ for FoE. We therefore compute $\partial o_e^{(l)}[i] / \partial r^{(l-1)}[i]$ (MoE) and $\partial o_e^{(l)}[i] / \partial r_{\text{full}}^{(l-1)}[i]$ (FoE). Let $J_e^{(l)} := \partial \text{FFN}_e^{(l)} / \partial u$ be the FFN Jacobian and $\mathbf{I} + A_{ii}^{(l)}$ the attention-plus-residual Jacobian, with $A_{h,ii}^{(l)}$ the analogue for group $h$'s own attention in FoE. Let $\gamma_{e'}^{(l-1)} := \partial w^{(l-1)}[i,e'] / \partial r_{\text{full}}^{(l-1)}[i] \in \mathbb{R}^d$ be the gate-weight gradient and $o_{e'}^{(l-1)}$ the corresponding expert output. Differentiating group $h$'s layer-$(l-1)$ update through both the experts and the router splits it into an FFN part and a routing part:
\[
  M_h^{(l-1)} = \mathbf{I} + \sum_{e' \in \mathcal{I}_h^{(l-1)}[i]} w^{(l-1)}[i,e']\, J_{e'}^{(l-1)},
  \qquad
  R_h^{(l-1)} = \sum_{e' \in \mathcal{I}_h^{(l-1)}[i]} o_{e'}^{(l-1)} \bigl(\gamma_{e'}^{(l-1)}\bigr)^{\!\top}.
\]
Here $M_h$ comes from differentiating the FFNs and $R_h$ is the rank-one outer-product term contributed by the gate weight's dependence on the router input. MoE is the $H = 1$ case, with $M$ and $R$ summed over all $k$ selected experts. We restrict attention to the token-diagonal term; the off-diagonal cross-token terms ($j \neq i$) carry through every chain below unchanged and are suppressed for readability.

\begin{proposition}
\label{prop:jacobian}
With the notation above, the per-expert inter-layer Jacobians of MoE and FoE are
\[
  \Bigl(\tfrac{\partial o_e^{(l)}[i]}{\partial r^{(l-1)}[i]}\Bigr)_{\!\text{MoE}}
    = J_e^{(l)}\bigl(\mathbf{I} + A_{ii}^{(l)}\bigr)\bigl(M^{(l-1)} + R^{(l-1)}\bigr),
\]
\[
  \Bigl(\tfrac{\partial o_e^{(l)}[i]}{\partial r_{\text{full}}^{(l-1)}[i]}\Bigr)_{\!\text{FoE}}
    = J_e^{(l)}\, \frac{1}{H}\sum_{h=0}^{H-1} \bigl(\mathbf{I} + A_{h,ii}^{(l)}\bigr)\bigl(M_h^{(l-1)} + R_h^{(l-1)}\bigr).
\]
That is, FoE replaces MoE's single block by a scale-preserving \emph{mean} of $H$ MoE-like blocks.
\end{proposition}

\paragraph{Proof sketch.}
The gradient chains through three stages: the previous-layer block $(M+R)$, the attention-plus-residual $(\mathbf{I}+A)$, and $\text{FFN}_e$. In MoE this is a single path. In FoE the $H$ groups each run their own attention but all read the same source $r_{\text{full}}^{(l-1)}[i]$, and the cross-group mean assembles expert $e$'s input as $r_{\text{full}}^{(l)}[i] = \frac{1}{H}\sum_h r^{(l)}[h,i]$, so $\partial r_{\text{full}}^{(l)}[i] / \partial x^{(l)}[h,i] = \frac{1}{H}(\mathbf{I} + A_{h,ii}^{(l)})$. Summing the chain rule over the $H$ groups that share the source yields the stated mean. \hfill$\square$

\paragraph{Why this preserves MoE's training behaviour.}
Three properties follow directly from the mean structure.
\begin{itemize}
  \item \textbf{Residual path preserved.} The leading term is $\frac{1}{H}\sum_h \mathbf{I} = \mathbf{I}$, the same identity (skip-connection) gradient path that keeps deep MoE models trainable.
  \item \textbf{Scale matched.} The block is a mean, not a sum, so its magnitude is $O(1)$ in $H$; the residual $1/H$ constant is absorbed by the following norm layer, which is scale-invariant. The gradient therefore neither explodes nor vanishes as $H$ grows.
  \item \textbf{Full coupling.} Every group enters the sum symmetrically, so all $H$ attentions and all $k$ previously selected experts feed every current expert, and cross-group gradient flows exactly as within-group gradient does. The routing term $R_h$ carries the global-softmax expert competition identically to MoE.
\end{itemize}

\paragraph{Two ablations show the design is deliberate.}
Table~\ref{tab:jacobian} places FoE alongside two variants that break this structure. Removing the cross-group mean makes the $H$ groups fully independent models, with cross-group gradient exactly zero; this is precisely the failure mode the concern imagines, and the mean is what avoids it. Mixing immediately after the experts, rather than deferring to the next layer's post-attention residual, pulls a single group's attention outside the average, leaving the router and every expert with only $1/H$ of the attention context. Deferring the mean until after attention is what puts all $H$ groups' attention into the gradient and gives routing the maximal cross-group information; it is also lossless in expressivity, since a linear mean reconstructs the same weighted sum over all $k$ experts, so no expert is dropped or reweighted.

The result is stated for arbitrary $H$, so nothing in it degrades as models grow. Because $H = n_{kv}$ and larger models carry more KV heads but also more experts, the per-group expert pool $E/H$ does not shrink pathologically with scale, and the mean structure above is unchanged by $H$. We therefore see no mechanism by which the gap to MoE would widen significantly at larger sizes. In fact, we expect the gap to shrink, because larger models have more experts and benefits diminish as the number of total experts grows, as MoE scaling laws show \citep{clark2022unified, ludz2024scalinglaws}. Our results already confirm this where the already small gap between MoE and FoE at 1B parameters shrinks in the 7B models we trained. In the FoE models we trained, each group only contains 8 experts, but with a larger model it could have as many as 64 experts. The difference between 64 and 8 experts (FoE group experts) is much larger than the difference between 64 and 512 (MoE total experts) thus we expect the gap to shrink as the model size grows.

\begin{table}[H]
  \centering
  \small
  \begin{tabular}{ll}
    \toprule
    Variant & Per-expert inter-layer Jacobian \\
    \midrule
    MoE
      & $J_e^{(l)}\bigl(\mathbf{I} + A_{ii}^{(l)}\bigr)\bigl(M^{(l-1)} + R^{(l-1)}\bigr)$ \\
    \textbf{FoE (ours)}
      & $J_e^{(l)}\, \frac{1}{H}\sum_{h}\bigl(\mathbf{I} + A_{h,ii}^{(l)}\bigr)\bigl(M_h^{(l-1)} + R_h^{(l-1)}\bigr)$ \\
    No cross-group mean
      & $J_e^{(l)}\bigl(\mathbf{I} + A_{h_e,ii}^{(l)}\bigr)\bigl(M_{h_e}^{(l-1)} + R_{h_e}^{(l-1)}\bigr)$, and $0$ for every other group \\
    Mean after experts
      & $J_e^{(l)}\bigl(\mathbf{I} + A_{h_e,ii}^{(l)}\bigr)\, \frac{1}{H}\sum_{h}\bigl(M_h^{(l-1)} + R_h^{(l-1)}\bigr)$ \\
    \bottomrule
  \end{tabular}
  \caption{Per-expert inter-layer Jacobian with the router differentiated, for MoE, FoE, and two FoE ablations. Group $h_e$ denotes expert $e$'s own group. Only FoE reproduces MoE's structure, as a scale-preserving mean of $H$ MoE-like blocks. Removing the cross-group mean decouples the groups, making the cross-group gradient exactly zero; taking the mean after the experts instead leaves a single group's attention outside the average, so routing sees only $1/H$ of the attention context.\label{tab:jacobian}}
\end{table}

\subsection{Topology Adaptability}
\label{sec:topology}

Section~\ref{sec:design} argues that FoE couples to the model, through $H = n_{kv}$, rather than to the hardware. Table~\ref{tab:topology-sweep} makes this concrete by evaluating the bandwidth-weighted per-layer ratio $T_{\text{moe}}/T_{\text{foe}}$ of Table~\ref{tab:comm-derivation-table} across deployment layouts, for a single $H = 8$ checkpoint with $k = 8$ and $r = 18$. The same weights are assumed throughout; only the mapping of expert groups onto GPUs changes.

Two observations follow. First, the ratio never drops below $1$, so FoE is never worse than MoE on any layout considered. Second, layouts with the same total GPU count but different shapes (for example $4 \times 8$ and $8 \times 4$, both $G = 32$) land at $5.91\times$ and $6.19\times$; a model trained for one is not mis-specialised for the other. For $G \leq H$ the FoE all-to-all vanishes entirely and the ratio is exactly $k$, which is why the small-$G$ entries saturate at $8.00\times$.

\begin{table}[H]
  \centering
  \small
  \begin{tabular}{lccc}
    \toprule
    & \multicolumn{3}{c}{GPUs per node ($G_n$)} \\
    \cmidrule(lr){2-4}
    Nodes ($N$) & $2$ & $4$ & $8$ \\
    \midrule
    $1$ & $8.00\times$ & $8.00\times$ & $8.00\times$ \\
    $2$ & $8.00\times$ & $8.00\times$ & $5.79\times$ \\
    $4$ & $8.00\times$ & $6.43\times$ & $5.91\times$ \\
    $8$ & $6.66\times$ & $6.19\times$ & $5.98\times$ \\
    \bottomrule
  \end{tabular}
  \caption{Predicted per-layer communication-time ratio $T_{\text{moe}}/T_{\text{foe}}$ for one $H = 8$ checkpoint deployed across node counts $N$ and node widths $G_n$ (total GPUs $G = G_n N$), with $k = 8$ and $r = 18$. Higher favours FoE. Values are analytic, obtained from the closed form in Table~\ref{tab:comm-derivation-table}.\label{tab:topology-sweep}}
\end{table}

\subsection{Expert-Group Balance During Training}
\label{sec:balance}

\begin{figure}[H]
    \centering
    \includegraphics[width=0.5\textwidth]{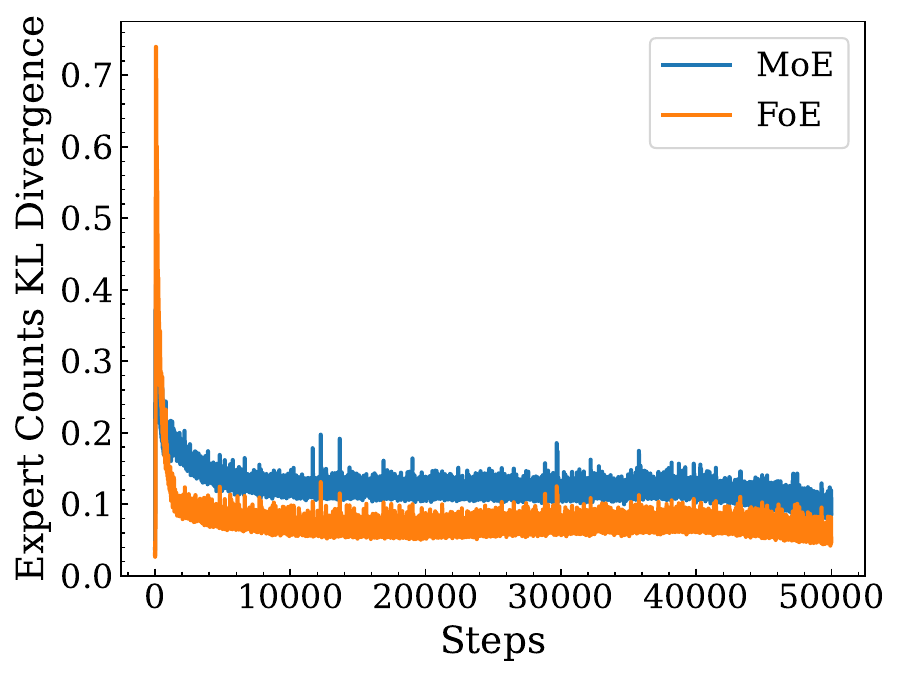}
    \caption{Expert-group balance across pretraining for the 7B models (lower is better). FoE maintains a more uniform expert-selection distribution than MoE throughout training, not only at convergence.\label{fig:balance-training}}
\end{figure}

Section~\ref{sec:lar_analysis} reports load balance for the converged model. Figure~\ref{fig:balance-training} shows the same comparison over the course of pretraining for the 7B models, using the per-batch KL divergence of the expert-selection distribution from uniform. Lower is better. FoE holds a lower value than its size-matched MoE baseline at every checkpoint, from the first logged step to the last, and does so with the load-balancing loss coefficient left at the baseline setting. Balance in FoE is thus a structural consequence of per-group top-$k/H$ routing rather than an artefact of tuning. It is worth separating the two levels at which balance arises. \emph{Between} groups, no tuning is necessary: per-group top-$k/H$ routing sends exactly $k/H$ selections from every token to every group at every layer, so all groups receive identical token counts by construction, even with the load-balancing loss coefficient set to zero. This is what yields the perfect $1.0$ load-balance rate of Section~\ref{sec:lar_analysis} on single-node, against the $\LoadDiscrepancyMoE\times$ max/median discrepancy of a standard MoE, which depends on its auxiliary loss. \emph{Within} a group, FoE reuses the standard MoE load-balancing loss unchanged, but applies it to a pool of $E/H$ experts rather than all $E$: a strictly easier balancing problem than the one MoE solves. Group balance is therefore controlled structurally rather than by tuning, and is less sensitive to hyperparameters than MoE, not more.

\subsection{Divisibility in Production MoE Models}
\label{sec:divisibility}

FoE sets $H = n_{kv}$ and requires the number of routed experts $E$, the top-$k$, and the query-head count $n_q$ each to be divisible by $H$. These are discrete structural constraints rather than continuous hyperparameters, and they are already satisfied by standard configurations: grouped-query attention alone forces $n_q$ to be divisible by $n_{kv}$. Table~\ref{tab:divisibility} surveys widely used open MoE models and reports the group count each would admit.

\begin{table}[H]
  \centering
  \small
  \setlength{\tabcolsep}{4pt}
  \begin{tabular}{lcccccc}
    \toprule
    Model & Total / Active & $n_q$ & $n_{kv}$ & Attention & $E$ / top-$k$ & Groups ($H$) \\
    \midrule
    OLMoE-1B-7B      & 7B / 1.3B   & 16  & 16    & MHA & 64 / 8  & 8 \\
    gpt-oss-120b     & 117B / 5.1B & 64  & 8     & GQA & 128 / 4 & 4 \\
    MiniMax-M2       & 230B / 9.8B & 48  & 8     & GQA & 256 / 8 & 8 \\
    GLM-4.5          & 355B / 32B  & 96  & 8     & GQA & 160 / 8 & 8 \\
    Qwen3-Coder-480B & 480B / 35B  & 96  & 8     & GQA & 160 / 8 & 8 \\
    Ling-1T          & 1T / 51B    & 64  & 8     & GQA & 256 / 8 & 8 \\
    DeepSeek-V3      & 671B / 37B  & 128 & 128$^{*}$ & MLA & 256 / 8 & 8$^{*}$ \\
    Kimi K2          & 1T / 32B    & 64  & 64$^{*}$  & MLA & 384 / 8 & 8$^{*}$ \\
    \bottomrule
  \end{tabular}
  \caption{Attention and MoE configurations of widely used open MoE models, with the expert-group count $H = n_{kv}$ each would admit under FoE. In almost every case $E$, top-$k$, and $n_q$ are divisible by the indicated $H$. $^{*}$These models use multi-head latent attention (MLA): K and V tensors are reconstructed at attention time from a compressed latent, thus FoE would be applied before/after the latent reconstruction on the KV hidden states. \label{tab:divisibility}}
\end{table}

\end{document}